\documentclass{article}
\usepackage{PRIMEarxiv}

\usepackage{booktabs}

\usepackage[utf8]{inputenc} 
\usepackage[T1]{fontenc}    
\usepackage[hidelinks]{hyperref}       
\usepackage{url}            
\usepackage{booktabs}       
\usepackage{nicefrac}       
\usepackage{microtype}      
\usepackage{lipsum}

\usepackage{amsmath,amssymb,amsfonts}
\usepackage{algorithmic}
\usepackage{graphicx}
\usepackage{textcomp}
\usepackage{xcolor}
\usepackage{multirow}
\usepackage{caption}
\usepackage{multirow}
\usepackage{arydshln}
\usepackage{caption}
\usepackage{subcaption}
\usepackage{wrapfig}

\usepackage[final]{changes}
\definechangesauthor[color=darkblue]{ss}
\definechangesauthor[color=darkorange]{a}

\definecolor{blue}{rgb}{0,0,0.8}
\definecolor{darkgreen}{rgb}{0.1,0.5,0.25}
\definecolor{darkblue}{rgb}{0.1,0.26,0.65}
\definecolor{darkorange}{rgb}{0.7,0.23,0.1}
\definecolor{codegreen}{rgb}{0,0.6,0}
\definecolor{codegray}{rgb}{0.5,0.5,0.5}
\definecolor{codepurple}{rgb}{0.58,0,0.82}
\definecolor{backcolour}{rgb}{0.95,0.95,0.95}

\usepackage{todonotes}
\setcommentmarkup{\todo[color={authorcolor!20},size=\scriptsize]{#3: #1}}

\newcommand{\note}[2][]{\added[#1,comment={#2}]{}}

\title{Clinical Risk Prediction Using Language models: Benefits and Considerations
}

\author{
  Angeela Acharya\textsuperscript{$\alpha$}, Sulabh Shrestha\textsuperscript{$\alpha$}, Anyi Chen\textsuperscript{$\beta$}, Joseph Conte\textsuperscript{$\beta$}, Sanja Avramovic\textsuperscript{$\alpha$}, \\ \textbf{Siddhartha Sikdar\textsuperscript{$\alpha$}, Antonios Anastasopoulos\textsuperscript{$\alpha$}, Sanmay Das\textsuperscript{$\alpha$}} \\
  \textsuperscript{$\alpha$} George Mason University \\
  \textsuperscript{$\beta$} Staten Island Performing Provider System  \\
  \texttt{\{aachary, sshres2, savramov, ssikdar, antonis, sanmay\}@gmu.edu}\\
  \texttt{\{Achen, jconte\}@statenislandpps.org
  }}
  
\begin{document}

\maketitle

\begin{abstract}
The utilization of Electronic Health Records (EHRs) for clinical risk prediction is on the rise. However, strict privacy regulations limit access to comprehensive health records, making it challenging to apply standard machine learning algorithms in practical real-world scenarios. Previous research has addressed this data limitation by incorporating medical ontologies and employing transfer learning methods. In this study, we investigate the potential of leveraging language models (LMs) as a means to incorporate supplementary domain knowledge for improving the performance of various EHR-based risk prediction tasks.  Unlike applying LMs to unstructured EHR data such as clinical notes, this study focuses on using textual descriptions within structured EHR to make predictions exclusively based on that information. We extensively compare against previous approaches across various data types and sizes.
We find that employing LMs to represent structured EHRs, such as diagnostic histories, leads to improved or at least comparable performance in diverse risk prediction tasks. Furthermore, LM-based approaches offer numerous advantages, including few-shot learning, the capability to handle previously unseen medical concepts, and adaptability to various medical vocabularies.  Nevertheless, we underscore, through various experiments, the importance of being cautious when employing such models, as concerns regarding the reliability of LMs persist.

\end{abstract}

\section{Introduction}
Electronic Health Records (EHRs) are comprehensive digital repositories containing a diverse range of healthcare information, such as patient diagnoses, prescriptions, insurance claims, and laboratory test results \cite{pendergrass_using_2018}. 
There has been growing interest in using EHRs for a variety of tasks, including 
clinical risk prediction \cite{risk-prediction}. The value proposition is that the availability of prior medical information in a standardized format could allow healthcare providers to anticipate future diagnoses, hospitalization, readmission, or a variety of related downstream outcomes.

A frequently proposed real-world use case of EHRs is to stratify specific subpopulations of interest in terms of their risks of future conditions; for example, stakeholders may wish to use local data to identify individuals at high risk of substance use disorder or diabetes in order to target appropriate interventions. Unfortunately, in many cases, such data is limited, making it difficult to apply conventional machine learning pipelines. Is it possible to produce high-quality risk scores on a subpopulation of interest by incorporating appropriate prior knowledge? 

Previous studies have explored various methods to achieve this. For instance, approaches like GRAM  \cite{gram} and GBERT \cite{gbert} utilized the hierarchical information within the International Classification of Diseases (ICD) \cite{icd-10} ontology to enhance their predictive capabilities. 
Likewise, MedBERT \cite{rasmy_med-bert_2021} and RareBERT \cite{prakash_rarebert_2021}  adopted a transfer learning approach, where they modified the BERT architecture \cite{devlin_bert_2019} for EHR, initially training a comprehensive model on a larger EHR dataset and subsequently fine-tuning it for the specific downstream tasks in question.

The objective of our study is to investigate how language models (LMs) can enhance performance in these types of risk prediction tasks, particularly by harnessing the rich information available in structured EHR data, such as diagnostic history. Given the divergence in vocabularies and data structure between structured EHR data and unstructured textual data, existing LMs designed for structured EHRs have typically been pre-trained from scratch \cite{rasmy_med-bert_2021, li_behrt_2020}. However, because of the limited availability of EHR data, models exclusively pre-trained on EHRs may not encompass the same breadth of knowledge as models pre-trained on larger text corpora. For instance, one key motivation for this study was to identify individuals at a high risk of developing opioid use disorder (OUD) or substance use disorder (SUD), a concern that is becoming increasingly prevalent, particularly in the United States \cite{economic_cost_OUD, gensyn}. There exists a substantial body of literature discussing the associated risks of these conditions \cite{county-level, McLellan2017-eu}, and having a model equipped with even a portion of this prior knowledge would offer significant advantages.

We propose two new methods (``Sent-e-Med'' and ``LLaMA2-EHR'') that aim to leverage this broader range of prior knowledge. 
Standard risk prediction tasks often revolve around a classification objective where we train a task-specific loss function on a customized dataset. Considering that most large language models (LLMs) \cite{touvron2023llama, palm, gpt3} are originally designed for text generation, their application in risk score prediction might seem counterintuitive. We nevertheless show that, when evaluated using standard metrics like area under the ROC or precision-recall curve, these models significantly outperform more conventional machine learning pipelines. However, we also demonstrate a particular kind of brittleness with respect to changes in the risk-prediction prompt, 
highlighting the need for extreme caution in moving forward with pipelines based on LLM predictions. We especially call for careful evaluation practices using meaningful validation sets and traditional evaluation metrics.

\section{Background}
Structured EHR data encompasses elements such as diagnoses, procedures, and laboratory results. Following a standardized format, they are represented by predefined vocabularies such as ICD-9, ICD-10  \cite{icd-10} for diagnoses, Current Procedural Terminology (CPT) \cite{cpt} for procedures, and RxNorm \cite{rxnorm} for medications. Each EHR data entry contains a patient's longitudinal hospital visit history, comprising various medical elements (as depicted in Fig. 1). These visits exhibit irregular patterns, characterized by variations in time intervals and the total number of visits among different patients.

\begin{figure}[!b]
    \centering
    \includegraphics[width=0.3\textwidth]{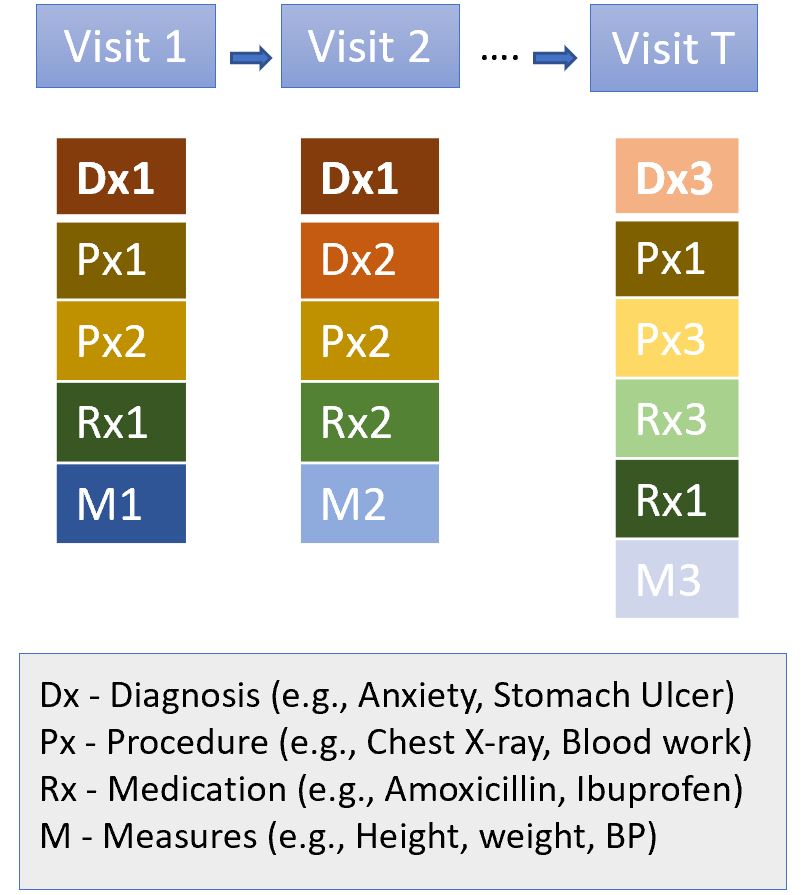}
    \caption{Representation of medical records for a single patient in a typical EHR: A visit may have a varying number of medical entities (i.e., diagnosis, procedure, medications, etc.)
    }
    \label{fig:EHR}
\end{figure}

The task of encoding a patient's medical history from EHRs to extract meaningful inferences remains an active area of research. For example, encoding categorical variables, such as diagnosis history, has evolved from traditional one-hot encoding \cite{opiod-prediction} to more sophisticated strategies, including the application of transformer-based methodologies \cite{ choi_doctor_2016, Zhang_2018}. Consequently, in order to encapsulate the information contained within EHRs, researchers have utilized a wide range of techniques, spanning from classical machine learning models like logistic regression and random forest \cite{DAI2015189,random-forest} to more recent deep learning models \cite{choi_doctor_2016, retain,  Barbieri2020BenchmarkingDL}.


Publicly available EHR sources are scarce and mostly focus on inpatient records \cite{mimiciv_v1, pollard_eicu_2018}, representing only a fraction of the healthcare landscape. A significant challenge in most of the previous EHR-based techniques is their requirement for large datasets. To overcome such challenges, recent studies enhance EHR representations by leveraging domain knowledge, such as the ICD code hierarchy \cite{icd-heirarchy}. For instance, GRAM \cite{gram} employed an attention-based mechanism--a widely adopted method across various applications \cite{retain, 10.1007/978-3-030-61255-9_14}--to identify meaningful connections among hierarchically organized medical concepts. A similar approach was taken by G-BERT \cite{gbert}, which employed graph neural networks for this purpose.

Recent studies have also incorporated methods based on LMs into various EHR tasks. For example, a variety of architectural variations based on BERT have been formulated \cite{ rasmy_med-bert_2021, prakash_rarebert_2021, li_behrt_2020}. BEHRT, a deep neural sequence model, has been introduced for the concurrent prediction of the likelihood of 301 different medical conditions in future patient visits \cite{li_behrt_2020}. MedBERT \cite{rasmy_med-bert_2021} has demonstrated its effectiveness in disease prediction studies, even when dealing with limited training datasets. Various other techniques, including RareBERT \cite{prakash_rarebert_2021} and CEHR-BERT \cite{cehr-bert}, have also been proposed in this context. The preference for BERT-based techniques is primarily due to their suitability for classification tasks, which aligns well with the nature of risk prediction in healthcare.
Despite their impressive performances, these methods are limited by their dependency on specific medical vocabularies, restricting their adaptability across different datasets and tasks. Even though these approaches utilize the framework of LMs, they fail to leverage the inherent knowledge embedded within these models.

Historically, clinical notes have typically been rich sources of textual information, and the application of LMs in the context of EHRs has predominantly focused on the analysis of clinical notes. This has given rise to models such as ClinicalBERT \cite{clinical-bert}, GatorTron \cite{yang2022gatortron}, each tailored to extract valuable insights from clinical narratives.

In this research, our goal is to explore the feasibility of representing structured EHR data, such as diagnostic history, as textual information. This approach offers greater versatility and independence from specific vocabularies, providing an opportunity to harness the existing knowledge embedded within LMs, among other benefits. In addition, we aim to conduct performance comparisons between LM based techniques and other contemporary approaches, such as MedBERT and GRAM, in contrast to more traditional methods like random forest and logistic regression.


\section{Experimental Setup} \label{sec:method}

\subsection{Problem Statement}
For simplicity, we consider the medical record of a single patient. A patient can have multiple visits in their medical history. Let $\mathbf{V} = [V_1, V_2, V_3,...,V_T]$ represent the list of all visits in chronological order. Our objective is to investigate whether past medical diagnoses can be informative in predicting the future risk of different conditions such as Substance Use Disorder (SUD), Opioid Use Disorder (OUD), and Diabetes.
Let $V_t = \{D_1, D_2, ..., D_N\}$, where $D_n$ is the $n^{th}$ diagnosis of the patient within visit $V_t$. 
It is important to note that the number of diagnoses within each visit can vary; for instance, $V_1 = \{D_1, D_2\}$ and $V_2 = \{D_1, D_2, D_3\}$. 
Furthermore, both the number of visits and the intervals between visits can also differ among patients. 
\deleted[id=ss]{For instance, the set of visits for patient 1 might look like this: $V_1 = \{\{D_1, D_2, D_3\}, \{D_1\}, \{D_1, D_2\},\{D_1, D_2, D_3, D_4\}\}$, whereas for patient 2 it might look like this: $V_2 = \{\{D_1\}, \{D_1, D_2\}\}$.} 

\subsection{Data}
We experiment using three datasets with differing sources and sizes. Our focus is on utilizing historical diagnoses information derived from patients' medical records. This diagnostic data is represented using either ICD-9 or ICD-10 codes, or SNOMED CT codes \cite{snomed_ct}.

\noindent\textbf{\em SIPPS.}
This is a de-identified dataset obtained from the Staten Island Patient Provider Service (SIPPS) network. 
It is community-specific and includes inpatient, emergency room (ER) visits, and outpatient records from two different hospitals (i.e., Staten Island University Hospital and Richmond Medical Center) representing the population of the Borough of Staten Island in New York. It contains records from the years 2019-2021. The diagnosis information in the SIPPS dataset is represented by ICD 10 codes.\footnote{Due to privacy and ethical considerations, the SIPPS data is not openly shared or published. However, qualified researchers who wish to access the data for legitimate research purposes may contact the corresponding author for data sharing inquiries.}

\noindent\textbf{\em MIMIC-IV.} \cite{johnson_alistair_mimic-iv_nodate}
This data is publicly available to use and is sourced from two in-hospital database systems: a custom hospital-wide EHR and an ICU-specific clinical information system. This includes de-identified medical records of patients who were admitted to the emergency department or one of the intensive care units of the Beth Israel Deaconess Medical Center (BIDMC) between 2008 to 2019. The diagnoses information in the MIMIC dataset is represented by both ICD 9 and ICD 10 codes.

\noindent\textbf{\em Synthea.} \cite{synthea}
This is an open-source package that generates realistic synthetic patients with associated health records for the top 10 reasons patients visit their primary care physicians and the top 10 chronic conditions responsible for years of life lost. We generated synthetic EHR data using Synthea and used it for different validation experiments. The diagnoses information in Synthea is represented by SNOMED CT codes which were later mapped to ICD 10 codes for easy analysis. 

\subsection{Data Processing} 
\begin{figure}[!thb]
    \centering
    \includegraphics[width=0.4\textwidth]{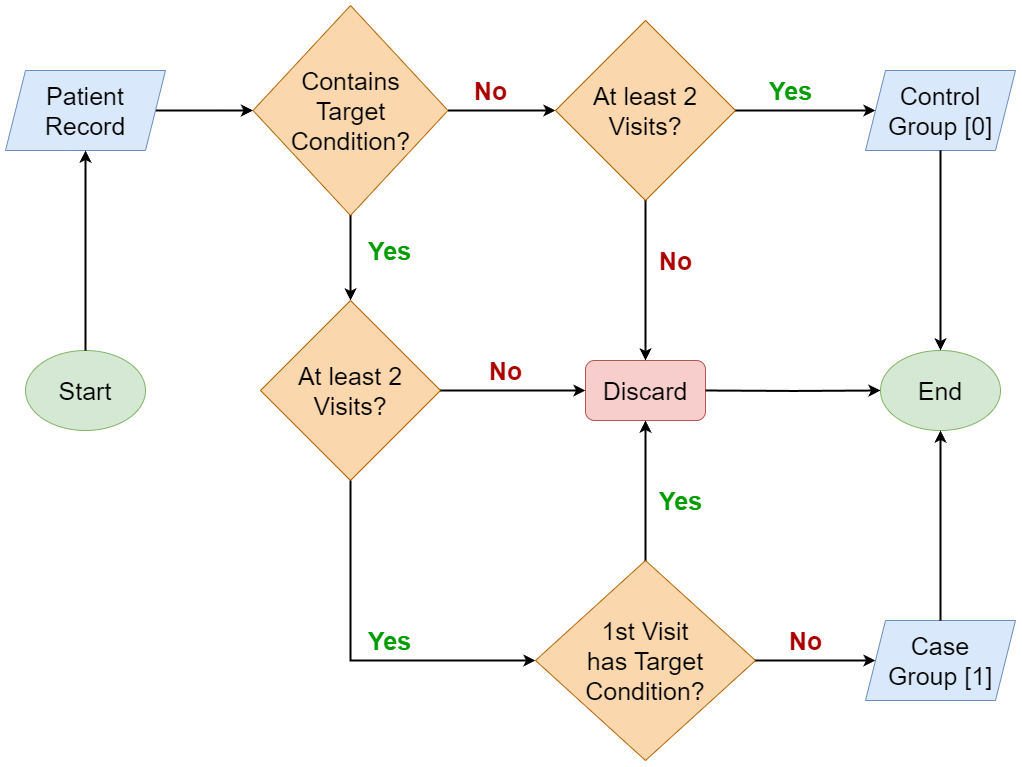}
    \caption{Process of creating patient groups. Patients that had at least one OUD/SUD/Diabetes diagnosis are put into Case Group while those that did not have any OUD/SUD/Diabetes diagnosis are put into Control Group.}
    \label{fig:patient-flowchart}
\end{figure}

For each risk prediction task, we create two groups of patients based on their diagnosis history. The process has been outlined in Fig. \ref{fig:patient-flowchart}. 

\noindent\textbf{Case Group.} 
The case group comprises patients who have at least one diagnosis that aligns with the prediction task. For instance, in the context of predicting SUD diagnoses, it includes any patient who has received at least one SUD diagnosis during their second visit or subsequent visits. Once a patient is selected, the target for the next visit prediction is the first visit in which the SUD diagnosis is documented.

\noindent\textbf{Control Group.} The control group is comprised of patients who do not have any diagnosis corresponding to the prediction task. For instance, when predicting SUD diagnoses, it includes patients with no SUD diagnosis documented in any of their visits. Once a patient is chosen, the target for the next visit prediction is the last visit in their medical history.

In both groups, all visits up to and including the target visit are kept while the latter visits are discarded.

\subsection{Selecting Data for Pretraining and Finetuning}
One major goal of this study is to explore different strategies for employing LMs in EHR analysis, which includes the option of either initializing the LM from scratch or fine-tuning an already existing LLM. Furthermore, we plan to assess how these approaches perform in comparison to alternative methods. To facilitate this investigation, it is essential to partition the dataset into sets designated for pretraining and fine-tuning.

For pretraining, we leverage the MIMIC-IV dataset, which has been widely employed in various studies \cite{gram, deep_oud, deep_ehr}. Our objective is to explore the potential of this publicly available dataset in learning generalizable representations that can be applied to multiple downstream tasks. The cohort in the dataset is diverse, encompassing over 114K patient records, and is not specific to any particular disease. The data is partitioned into training, test, and validation sets in a ratio of 7:2:1, respectively. To prevent data leakage, we ensure that the records of each patient are exclusively included in one of the sets. As a result, the pretraining set consists of approximately 80K patients. While the validation and test sets are not utilized during the pretraining phase, they are reserved for evaluation during the finetuning stage. This ensures that the data used for testing and validation are distinct from the data used in pretraining, avoiding any overlap.

We utilize four different types of data for the finetuning stage:
\begin{itemize}
\item MIMIC: Obtained from the original MIMIC cohort but is condition-specific (i.e., separate data for SUD, OUD, and Diabetes risk prediction).
\item MIMIC-small: Several smaller samples were chosen from the MIMIC dataset to explore how different models perform with limited data. 
\item SIPPS: As mentioned above, this is a private EHR data sourced from members within the SIPPS network.

\item Synthea-small: As with MIMIC-small, we extract several smaller samples from the data generated using the synthea framework.
\end{itemize}

Tables \ref{tab:pretrain} and \ref{tab:descr} provide a summary of the characteristics of the different patients in the pretraining and the finetuning datasets respectively.

\begin{table*}[!t]
\footnotesize
\begin{center}
\caption{Descriptive summary of the pretraining dataset}
\label{tab:pretrain}
\begin{tabular}{cc}
\toprule
\textbf{Characteristic} & \textbf{Pretraining data} \\
\toprule
Total number of patients & 84431 \\
Average Number of visits per patient & 3 \\
Average Number of codes per patient & 23\\
Vocabulary size& 27052\\
\bottomrule
\end{tabular}
\end{center}
\end{table*}

\begin{table*}[!t]
\footnotesize
\begin{center}
\caption{Descriptive summary of the finetuning datasets. The statistics are derived from the data post the processing stage.}
\label{tab:descr}
\begin{tabular}{cccccc}
\toprule
 & & \multicolumn{3}{c}{\textbf{Finetuning Data}} \\
\multirow{-2}{*}{\textbf{Outcome}} & \multirow{-2}{*}{\textbf{Characteristic}} &\textbf{ MIMIC} & \textbf{MIMIC-small} & \textbf{SIPPS} & \textbf{Synthea-small}\\
\toprule
 &Total number of patients & 77341  &  2650& 2182 &2650\\
 &Number of patients with event & 6984 & 750& 619 & 750\\
 &Average number of visits per patient &4&4&22  & 14\\
 &Average number of codes per patient &28&29&30 & 12\\
 \multirow{-5}{*}{SUD}&Vocabulary size &20196&6704&5556 & 189\\
 \midrule
 &Total number of patients & 83354  & 2780 &  2364 & 1743\\
 &Number of patients with event & 1585 & 130 & 110 & 83\\
 &Average number of visits per patient &4&4&28 & 16\\
 &Average number of codes per patient &30&29&39 & 12\\
 \multirow{-5}{*}{OUD}&Vocabulary size &21667&6800&6856 & 186\\
\midrule
 &Total number of patients & 34295  & 2182 &  2192 & 2182\\
 &Number of patients with event & 2201 & 619 & 459 & 619\\
 &Average number of visits per patient &4&4&22 & 10\\
 &Average number of codes per patient &25&29&31 & 7\\
 \multirow{-5}{*}{Diabetes}&Vocabulary size &16670&5933&5790 & 151\\
\bottomrule
\end{tabular}
\end{center}
\end{table*}

\subsection{Case Definition}
The experimental datasets employed in this study encompass a vocabulary comprising both ICD-9 and ICD-10 codes. To identify SUD, OUD, and Diabetes from the diagnosis history, the corresponding codes listed in Table \ref{tab:icdcodes} were utilized.

\begin{table}[h]
\footnotesize
\centering
\caption{ICD 9 and ICD 10 codes used for identifying OUD, SUD, and Diabetes in the datasets.}
\label{tab:icdcodes}
\begin{tabular}{lll}
\toprule
\textbf{Diagnosis} & \textbf{ICD 10 Codes} & \textbf{ICD 9 Codes} \\
\midrule
OUD & F11 & 304.00 - 304.03, 304.70 - 304.73, \\
&& 305.50 - 305.53 \\
\midrule
SUD & F10 - F19 & 291 - 292, 303.00 - 303.03, \\
&& 303.90 - 303.93, 304.00 - 304.03, \\
&& 304.10 - 304.13, 304.20 - 304.23, \\
&& 304.30 - 304.33, 304.40 - 304.43, \\
&& 304.50 - 304.53, 304.60 - 304.63, \\
&& 304.70 - 304.73, 304.80 - 304.83, \\
&&304.90 - 304.93, 648.30 - 648.34 \\
\midrule
Diabetes & E08 - E13 & 250.xx \\
\bottomrule
\end{tabular}
\end{table}

\section{Methods} \label{sec:current-approach}
We first describe how we adapted some previous methods and then outline our two proposed approaches.

\subsection{Previous Methods}

\subsubsection{MedBERT}
Proposed by Rasmy et al \cite{rasmy_med-bert_2021}, Med-BERT is an adaptation of the original BERT framework on structured EHR and was trained on diagnoses data coded using the ICD codes. While the pre-trained model and the original data were not available, we made use of the framework and pre-trained it using the MIMIC data. This approach serves as a baseline for BERT-based EHR pretrained models and can aid in assessing the impact of pretraining on performance across various contexts. MedBERT was chosen because of its popularity, ease of implementation, and performance. For more details on the model please refer to the paper \cite{rasmy_med-bert_2021}.

\subsubsection{GRAM}
Proposed by Choi et. al \cite{gram}, this model supplements structured EHR with hierarchical information inherent to medical ontologies. The implementation of this model allows for the assessment of the effectiveness of a domain knowledge integration-based approach and a general deep learning model, without pre-training, in handling various data types.
\subsubsection{Logistic Regression \& Random Forest}
In many healthcare applications, there is a preference for simpler and interpretable models over complex ones, especially when dealing with limited datasets. We implemented a Balanced Random Forest model and a Logistic Regression model to test simpler models. Since our tasks demand working with heavily imbalanced data, a balanced model was implemented.  Based on experiments involving different features, we selected the top 10\% of the most frequent diagnosis codes among patients with the target condition as predictors. To represent the presence of each selected diagnosis code, we employed multi-hot encoding, indicating whether the corresponding diagnoses were present in the patient's medical history.


\begin{figure*}[t]
\centering
\includegraphics[width=0.7\textwidth]{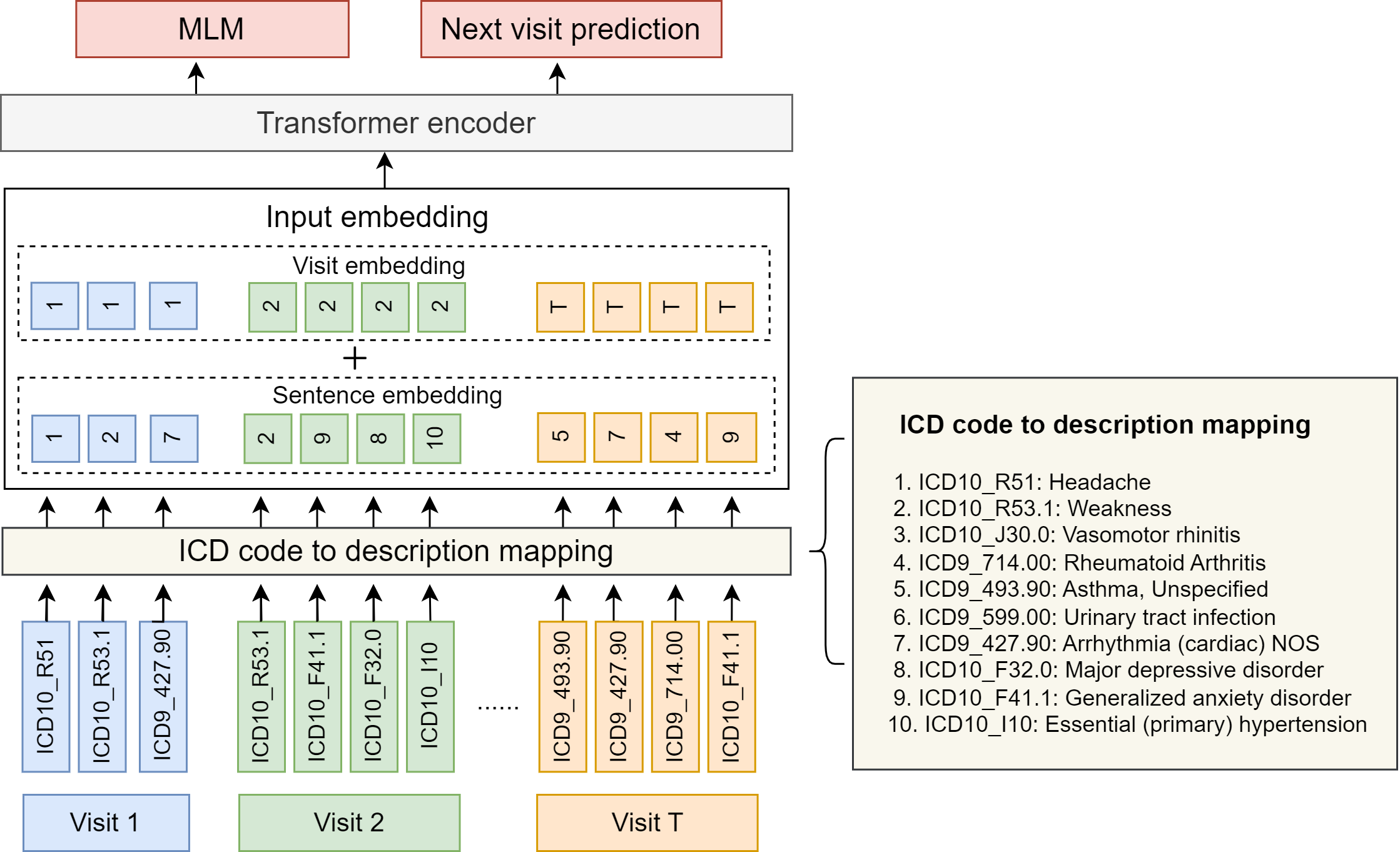}
\caption{High-level overview of the Sent-e-Med architecture: For each medical code, sentence embeddings and visit embeddings are extracted and subsequently combined before being fed into the transformer encoder as input.}
\label{fig:approach}
\end{figure*}

\subsection{Sent-e-Med}
Sent-e-Med is a direct modification of MedBERT. 
Any other LM designed explicitly for structured EHR could have been viable as well. 
As depicted in Fig. \ref{fig:approach}, we represent each visit as a variable-length sequence of diagnosis codes, which are then mapped to textual descriptions and encoded using S-BERT \cite{reimers_sentence-bert_2019} to create ``sentence embeddings''. The sentence embeddings, derived from pre-trained S-BERT, remain unchanged throughout the process to preserve their already rich knowledge obtained through text corpora. We also tried not freezing the sentence embeddings but did not obtain satisfactory results.
 Additionally, ``visit embeddings'' serve as unique identifiers for each visit, akin to segment embeddings in BERT and are randomly initialized and updated during pre-training.

Similar to MedBERT, we do not employ [CLS] and [SEP] tokens and omit positional encodings due to format differences between EHR data and standard text. The final input embedding is generated by summing the sentence and visit embeddings and is passed to the transformer encoder.

Once the model is pre-trained, a classification layer (fully-connected layer) is added on top of the model to make it suitable for finetuning on the risk prediction tasks. 

Sent-e-Med \replaced[id=ss]{makes modifications to}{handles the limitations of} MedBERT in three ways outlined below:

\subsubsection{Representations from Textual Descriptions of Medical Codes}
MedBERT and similar BERT-based approaches for structured EHRs \cite{prakash_rarebert_2021, li_behrt_2020}) start with random embeddings for medical code tokens and then learn these embeddings from scratch during pretraining. However, this approach can take a long time to converge and requires substantial data for learning generalizable representations, which EHR lacks compared to natural language. The same issue arises when novel medical codes are encountered during fine-tuning.

To \replaced[id=ss]{overcome}{solve} these limitations, we leverage the textual description of medical codes and initialize the medical code tokens with embeddings obtained from a sentence encoder. This encoder encodes input sentence(s) to produce a single embedding that preserves the semantic relationship between sentences such that two sentences that are similar in \replaced[id=ss]{meaning}{descriptions} are encoded close to each other in the \replaced[id=ss]{embedding}{vector} space as measured by cosine similarity. SBERT \cite{reimers_sentence-bert_2019} is an example of such a sentence encoder that we use in this study and is based on finetuning BERT using Siamese networks and Triplet loss \cite{ferrari_triplet_2018}. A pre-trained SBERT model is used to encode the medical codes based on their textual descriptions (which are sentences). For instance, the textual description corresponding to the ICD 10 code ``F10.13'' is ``Alcohol abuse, with withdrawal'' which is input to SBERT to obtain a single embedding corresponding to the sentence which we use to initialize embedding for the medical code.  
We also experimented with Universal Sentence Encoder \cite{cer_universal_2018} but we found SBERT representations to work better in preliminary experiments. 
By adopting this pretraining approach, we overcome the aforementioned limitations, leading to improved performance and accelerated training.  

\subsubsection{Pretraining Objectives}
 MedBERT used two different pretraining objectives: prolonged length of stay (LOS) and masked language modeling (MLM). The MLM objective in MedBERT is motivated by BERT and involves masking 15\% of the medical codes and predicting the existence of those codes given their context {\em i.e.} the remaining 85\% of the medical codes. The prolonged LOS objective is specific to MedBERT and corresponds to a binary classification task \replaced[id=ss]{of predicting}{where the outcome is 0 or 1 based on } whether a patient had to stay in the hospital for more than 7 days in any of their EHR visit sequences. While this objective is useful for certain kinds of problems, it is not very useful for risk prediction tasks. \deleted[id=ss]{Thus, using LOS would only increase the training time while not guaranteeing a good performance in the target risk prediction task.} \note[id=ss]{This cannot be said because there is not experiment trying this objective} Therefore, we introduce \textit{next visit prediction} as a new pretraining objective. Specifically, for each patient, we use all of the past diagnosis codes to predict the diagnosis codes in the next visit. Inspired by RoBERTa \cite{roberta}, we also experiment with a variant that only uses the MLM training objective.

\subsubsection{Freezing the Sentence Embeddings}
\begin{figure}[!t]
\centering
\includegraphics[width=0.4\textwidth]{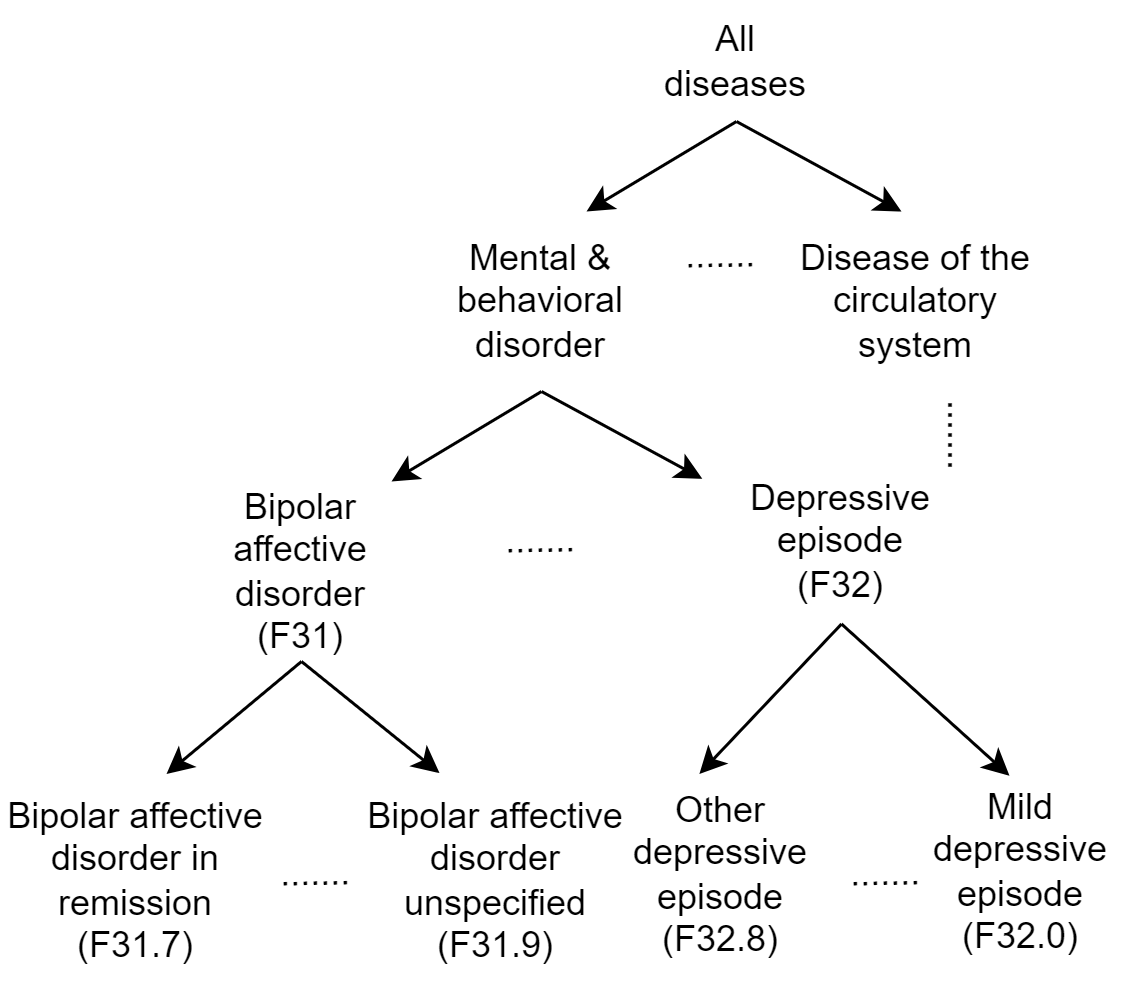}
\caption{An example demonstrating the hierarchical structure of ICD 10 diagnosis codes}
\label{fig:heirarchy}
\end{figure}

We keep the encoded sentence embeddings frozen during both pretraining and fine-tuning. This approach leverages pre-trained representations containing valuable knowledge about diagnostic associations, reducing training time and parameters. 
This approach offers an additional advantage: it's especially beneficial for medical concepts, such as those in the ICD hierarchy (see Fig. \ref{fig:heirarchy}), where related concepts share similar descriptions. By freezing sentence embeddings in Sent-e-Med, similar child concepts have similar embeddings even after training, enabling the model to deduce relationships between new codes and existing medical codes based on learned associations from related or similar codes in the training data. This indirect transfer of hierarchical information is more convenient than explicitly providing tree-based information, as seen in approaches like GRAM \cite{gram} or G-BERT \cite{gbert}.

\begin{figure*}[!t]
\centering
\begin{subfigure}{.5\textwidth}
\centering  

\fbox{\includegraphics[width=0.89\linewidth]{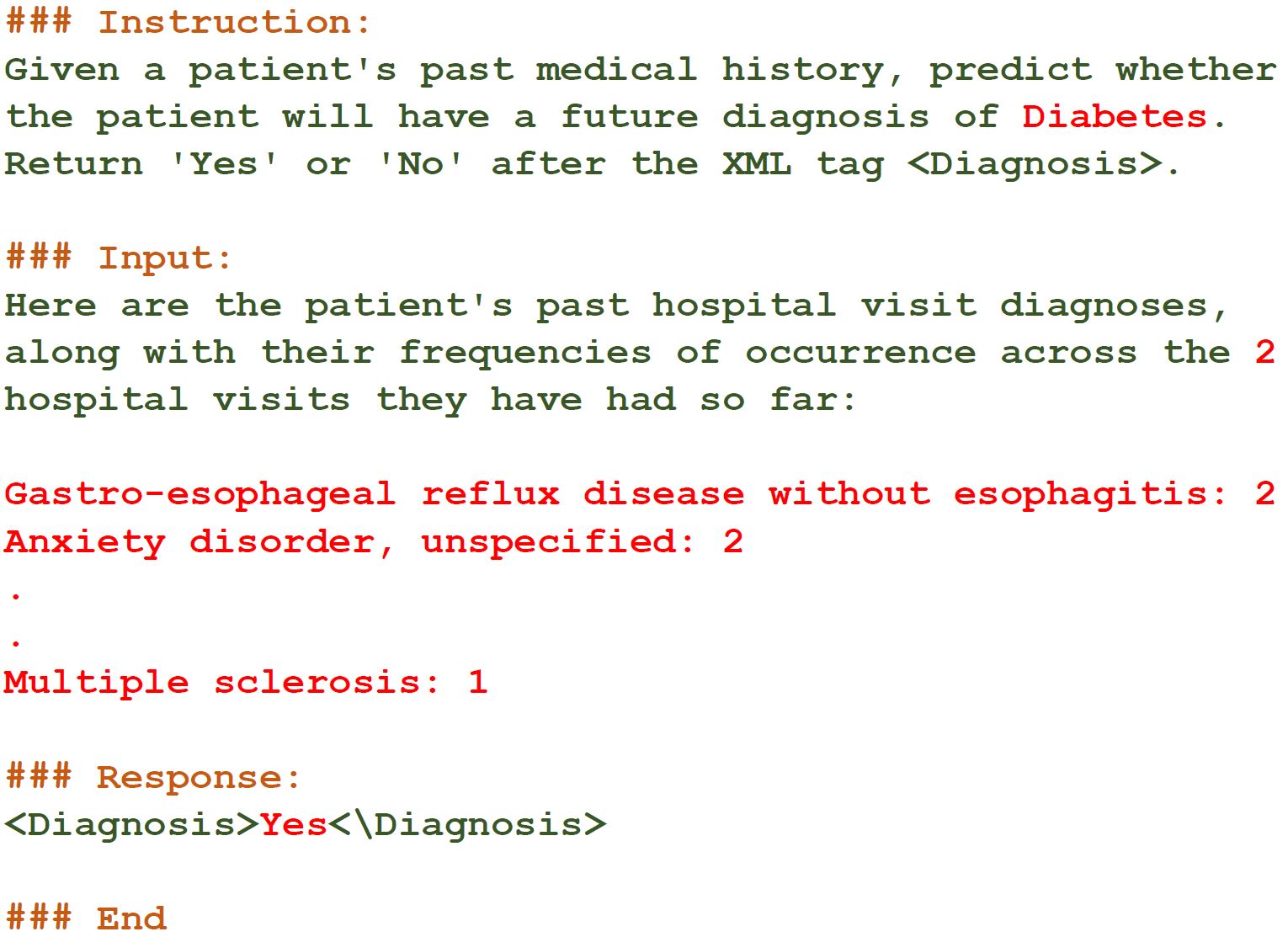}}
  \caption{Prompt 1}
  \label{fig:prompt1}
\end{subfigure}%
\begin{subfigure}{.5\textwidth}
  \centering
  \fbox{\includegraphics[width=1\linewidth]{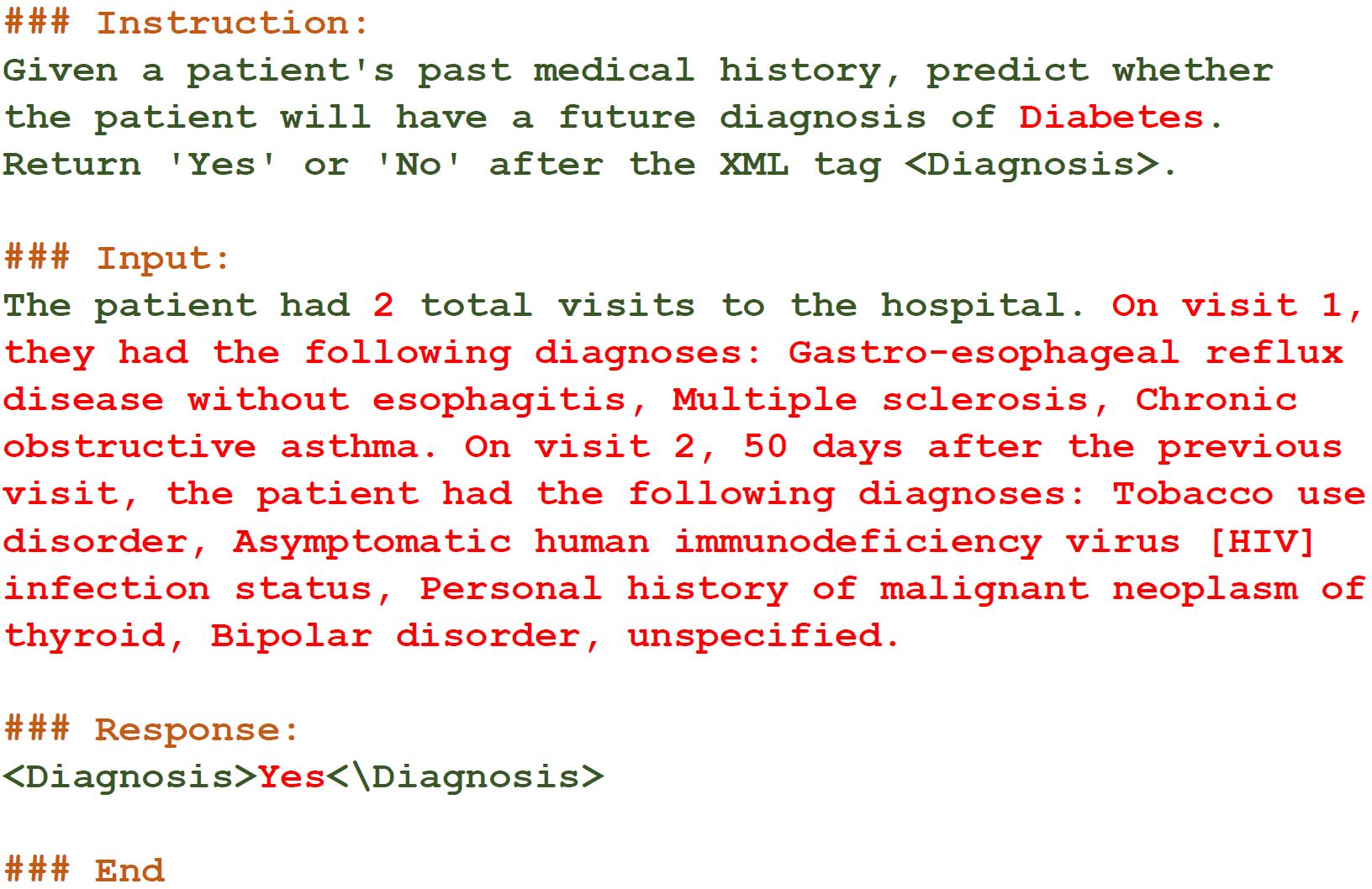}}
  \caption{Prompt 2}
  \label{fig:prompt2}
\end{subfigure}
\caption{Illustration of two distinct prompts employed in the fine-tuning of the LLaMA2-EHR model. Prompt 1 aggregates the frequency of diagnosis occurrences across multiple visits, while Prompt 2 evaluates diagnoses on a per-visit basis and incorporates information about the intervals between visits. Red highlights in the text are employed to indicate patient-specific variations in the information. \\
\textit{Note: Inputs in the prompts and the responses are just hypothetical examples.}}
\label{fig:prompts}
\end{figure*}

Note that we also conducted experiments with a version where the sentence embeddings were not frozen, but we did not obtain satisfactory results.

\subsection{LLaMA2-EHR}
We finetune the 7B-chat version of the LLaMA2 \cite{touvron2023llama} model, one of the top open-source LLMs \cite{LLM}, on structured EHR and dub it LLaMA2-EHR.\footnote{There's a larger, better-performing version (70B) available, however it is really difficult to use due to limited computational resources.}
To facilitate a more straightforward comparison with the other approaches, we fine-tune the model exclusively on diagnosis history. 

As mentioned earlier, the structure of our data differs from the original datasets used for training LMs, as it consists of an irregular sequence of patient visits. We explore two distinct approaches for utilizing this information and constructing input data for LLaMA2-EHR. One approach involves cumulatively aggregating all visit information to provide a summary of a patient's history (referred to as Prompt 1, as illustrated in Fig \ref{fig:prompt1}). The other approach utilizes temporal information to represent patient visits (referred to as Prompt 2, as shown in Fig \ref{fig:prompt2}). 

Unlike Sent-e-Med, which requires pretraining on the EHR corpus while only keeping the sentence embeddings frozen, LLaMA2-EHR focuses solely on fine-tuning for a limited number of epochs.

\subsection{Benefits of Text-based Representation}
Both Sent-e-Med and LLaMA2-EHR operate without dependency on specific coding systems. In the healthcare domain, different organizations may employ distinct vocabularies to encode medical entities, such as SNOMED CT or ICD-10 for diagnosis. Even within the same organization, the coding system undergoes frequent changes, e.g., transitioning from ICD-9 to ICD-10. By operating on textual descriptions, these models mitigate concerns about the specific vocabularies present in the data, enabling its application across diverse datasets.


\section{Results}
\label{sec:results}
\subsection{Evaluation}
Considering our risk prediction tasks, our main emphasis was on evaluating outputs in the form of probability scores rather than simple binary labels of 0 and 1. Thus, to gain deeper insights into the models' effectiveness in distinguishing patients based on varying risk probabilities, we selected Area under the ROC Curve (ROC AUC) and Precision-Recall AUC (PR AUC) as our preferred evaluation metrics.

While most of our experimental models can be straightforwardly evaluated by extracting output probabilities and employing these chosen metrics, assessing LLaMA2 for classification is notably more challenging because it often outputs verbose text that needs to be post-processed to be mapped to a certain class. To address this, we use the input up to the XML tag \texttt{<Diagnosis>} and obtain the probability distribution of the possible next tokens. Ideally, this stage should yield high probabilities for tokens ``Yes'' or ``No''. However, another challenge arises from the multiple representations of ``Yes'' and ``No'' (e.g., ``No'', ``no'', ``NO''). To tackle this, we take the tokens with the top-$k$ probability scores (we used $k=20$), sum those associated with potential representations of ``Yes'' and ``No'' separately, and then apply a softmax over the two values. This allows us to obtain estimated probabilities for ``Yes'' and ``No'',  enabling us to apply the chosen metrics.

\subsection{Comparative Analysis}
\label{sec:results-table}
\added[id=ss]{As outlined in Section \ref{sec:method},} we experimented with four different types of data for three separate tasks during finetuning: OUD, SUD, and Diabetes risk prediction. We used the same train, test, and validation sets for all these models \replaced[id=ss]{for fair comparison}{make the comparison more accurate}. The validation and test sets used during the finetuning stage in the BERT-based models are never used during pretraining. Table \ref{tab:results} summarizes the results we obtained on our test sets. The results depict the average performances across 5 different runs.

We assess the significance of performance variations between the proposed approaches and the top-performing baseline by employing a bootstrapping method. This method includes 1000 iterations of random sampling with replacement from the prediction values, calculating ROC AUC and PR AUC values for each sample. Subsequently, we determine the frequency with which the AUC values of the proposed approaches surpass those of the top baseline (or how often the top baseline performance lags behind that of the proposed methods). The $p$-value is then computed based on this information.

We omit the results obtained from using Prompt 2 on LLaMA2-EHR due to its suboptimal performance on SIPPS. The LLaMA2 \cite{touvron2023llama} model's sequence length is limited to 4096 tokens which cannot accommodate longer input sequence resulting from high number of visits in SIPPS (see Table \ref{tab:descr}). For other datasets, the results closely resembled those obtained from Prompt 1.

The following are the main takeaways from the results shown in Table \ref{tab:results}:
\begin{itemize}
\item Among the baseline models, MedBERT appears to exhibit the best ROC AUC and PR AUC performances in the majority of the tested tasks and datasets.
\item The pretraining-based methods (i.e., MedBERT, Sent-e-Med, and LLaMA2-EHR) are significantly better than other methods in terms of both metrics in all but one dataset (SIPPS). In the SIPPS dataset, all of the experimented models had almost similar performances, except for the diabetes risk prediction task. 
\item In all the risk prediction tasks and datasets we explored, LLaMA2-EHR consistently demonstrated superior (or at least equal) AUC ROC and PR AUC performances compared to the other models. This is particularly noteworthy when we focus on the PR AUC values. For example, in the OUD prediction task, LLaMA2-EHR achieved a performance of 72\%, while MedBERT scored 57\% on the MIMIC-small dataset. Similarly, on the Synthea-small dataset, LLaMA2-EHR obtained a PR AUC of 84\%, surpassing the 60\% achieved by MedBERT.
\item Overall, the models that utilized the textual inputs (Sent-e-Med and LLaMA2-EHR) had better performances as compared to others.
\item Between the two variants of Sent-e-Med, both demonstrated nearly identical performance, although the variant that only used the MLM training objective performed slightly better overall. It is possible that introducing the next visit prediction objective added more complexity to the model making it harder to learn effective representations.  Consequently, we just present results from the model utilizing only the MLM objective.
\item In most cases, simpler models such as logistic regression and random forest did not exhibit strong performance, particularly when dealing with larger datasets like MIMIC. 

Section \ref{sec:discussion} delves into the potential explanations for these observations and explores additional noteworthy findings.

\end{itemize}

\begin{table}[t]
\small
\centering
\caption{Average ROC AUC and PR AUC values across our datasets in the context of our prediction tasks.  BR represents the base rate, i.e., the ratio of positive to negative samples. These results pertain to the test set of the fine-tuning datasets.\\
\small \textit{Note: For Sent-e-Med and LLaMA2-EHR, values marked with an asterisk indicate a significant deviation from the best baseline performance.\\
*p<0.1, **p<0.05, ***p<0.01 }}
\label{tab:results}
 \setlength{\tabcolsep}{3.5pt}
\resizebox{1\textwidth}{!}{
\begin{tabular}{clllllllllllll}
\toprule
 &  & \multicolumn{3}{c}{\textbf{MIMIC}} & \multicolumn{3}{c}{\textbf{MIMIC-small}} & \multicolumn{3}{c}{\textbf{SIPPS}} & \multicolumn{3}{c}{\textbf{Synthea-small}}\\
\multirow{-2}{*}{\textbf{Outcome}} & \multirow{-2}{*}{\textbf{Model}} & \textbf{ROC AUC} & \textbf{PR AUC} & \textbf{BR} & \textbf{ROC AUC} &\textbf{PR AUC}&\textbf{BR} & \textbf{ROC AUC} & \textbf{PR AUC} & \textbf{BR} & \textbf{ROC AUC} & \textbf{PR AUC} & \textbf{BR}\\
\toprule
 & Random Forest & 0.81 & 0.34 &0.1 & 0.85 & 0.67 & 0.4& 0.74 &0.57 & 0.4&0.77 & 0.59 &0.4\\
 & Logistic                     & 0.81 & 0.32&0.1 & 0.75 & 0.62 & 0.4&0.74 & 0.57 & 0.4&0.79 & 0.63&0.4\\
  & GRAM                         & 0.85 & 0.35 & 0.1&0.87 & 0.65 &0.4& 0.74 & 0.54 &0.4& 0.86 & 0.73&0.4\\
& MedBERT                      & 0.89 & 0.70 &0.1 &0.87 & 0.77 & 0.4 & 0.75 & 0.57 &0.4& 0.87 & 0.75&0.4\\ 
   \cmidrule(lr){2-2} 
      \cmidrule(lr){3-14}
 & Sent-e-Med              & $0.91^{***}$ & $\textbf{0.73}^{***}$ &0.1& $0.89^{*}$ & 0.79 & 0.4 &\textbf{0.76} & \textbf{0.58} & 0.4&0.87 & 0.76&0.4\\
\multirow{-6}{*}{SUD} & LLaMA2-EHR & $\textbf{0.92}^{***}$  & $\textbf{0.73}^{***}$ &0.1& $\textbf{0.93}^{***}$ & $\textbf{0.89}^{***}$ & 0.4 &0.74 & 0.56 & 0.4& $\textbf{0.88}^{**}$& $\textbf{0.78}^{**}$&0.4\\
 \midrule
 & Random Forest & 0.87    & 0.18 & 0.02& 0.71 & 0.34 & 0.05&\textbf{0.75} & 0.17 & 0.05&0.88 & 0.33&0.05\\
 & Logistic                         & 0.86 & 0.19 &0.02& 0.83 & 0.39 &0.05 & \textbf{0.75} & 0.17 & 0.05&0.88 &0.33&0.05 \\
 & GRAM                             & 0.90 & 0.22 &0.02& 0.87 & 0.30 & 0.05& 0.74 & 0.17 & 0.05&0.92 & 0.57&0.05 \\
  & MedBERT                          & 0.91 & 0.53 &0.02& 0.89 & 0.57 & 0.05& 0.74 & 0.19 & 0.05&0.93 & 0.6&0.05 \\
   \cmidrule(lr){2-2} 
      \cmidrule(lr){3-14}
 & Sent-e-Med                  & $\textbf{0.95}^{***}$ & $0.58^{*}$ & 0.02& 0.93 & 0.57 & 0.05 &\textbf{0.75} & 0.19 & 0.05&0.95 &0.7&0.05 \\
\multirow{-6}{*}{OUD}& LLaMA2-EHR & $\textbf{0.95}^{***}$ & $\textbf{0.59}^{***}$ & 0.02& $\textbf{0.95}^{**}$ & $\textbf{0.72}^{***}$ &0.05 & 0.73 &\textbf{0.2} &0.05 & $\textbf{0.98}^{***}$ &
$\textbf{0.84}^{***}$&0.05 \\
\midrule
& Random Forest & 0.75 & 0.22  &0.07& 0.75 & 0.58 &0.4& 0.82 & 0.59  &0.27&0.88 & 0.74 &0.4\\
 & Logistic &0.72 &0.19  &0.07& 0.74 & 0.5&0.4& 0.81 &0.59 &0.27&0.88 &0.76&0.4\\
 & GRAM &0.77  & 0.16 &0.07&0.71 & 0.4 &0.4& 0.71& 0.52 &0.27&0.93 & 0.86&0.4 \\
 & MedBERT & 0.84 & 0.61 & 0.07&0.8& 0.72 &0.4&0.79 &0.57  &0.27& \textbf{0.94} &0.86&0.4 \\
   \cmidrule(lr){2-2} 
      \cmidrule(lr){3-14} 
 &Sent-e-Med & $\textbf{0.89}^{***}$& $\textbf{0.67}^{*}$ &0.07&$\textbf{0.83}^{**}$&$0.74^{***}$ &0.4&$0.81^{***}$ &$0.62^{***}$ &0.27&\textbf{0.94} &$\textbf{0.88}^{**}$&0.4\\
\multirow{-6}{*}{Diabetes } & LLaMA2-EHR & $0.88^{***}$ &$\textbf{0.67}^{*}$& 0.07&$\textbf{0.83}^{*}$ &$\textbf{0.76 }^{**} $&0.4& $\textbf{0.83}^{***}$& $\textbf{0.64 }^{***}$& 0.27&\textbf{0.94}&$\textbf{0.88}^{*}$&0.4\\
\bottomrule

\end{tabular}}
\end{table}

\subsection{LLaMA2-EHR Sensitivity Analysis}
\subsubsection{Changing the Instructions}
Our results have indicated that LLaMA2-EHR generally performs well when tested with the same instructions used during the fine-tuning stage. However, we were also interested in exploring what happens when we make slight alterations to the instructions. For instance, what if we modify the instruction to predict the risk (High/Low) of having a diagnosis, as opposed to asking a Yes/No question?

We tested two different models. The first model, finetuned on the MIMIC dataset, consistently failed to adhere to the provided instructions, outputting ``Yes'' or ``No'' rather than ``High'' or ``Low'' in all instances. The second model, finetuned on the MIMIC-small dataset, exhibited the desired behavior, predicting ``High'' and ``Low'' as instructed in most cases. However,  in approximately 35\% of the instances, the outputs were inverted, meaning it predicted ``High'' when the output with the original prompt was ``No'' and ``Low'' when the output was ``Yes''.

We also conducted this experiment using the original LLaMA2-7B-chat model and found that the model could generally adhere to and understand the modified instruction, yielding conforming probabilities of High/Yes and Low/No. 

This discrepancy raises concerns about the adaptability of the model to changing instructions. It suggests that fine-tuning on our datasets might have induced catastrophic forgetting \cite{catastrophic} particularly when dealing with a large volume of data.

\subsubsection{Changing the Inputs}
\begin{figure*}[!t]
\centering
\begin{subfigure}{.5\textwidth}
\centering  \fbox{\includegraphics[width=0.95\linewidth]{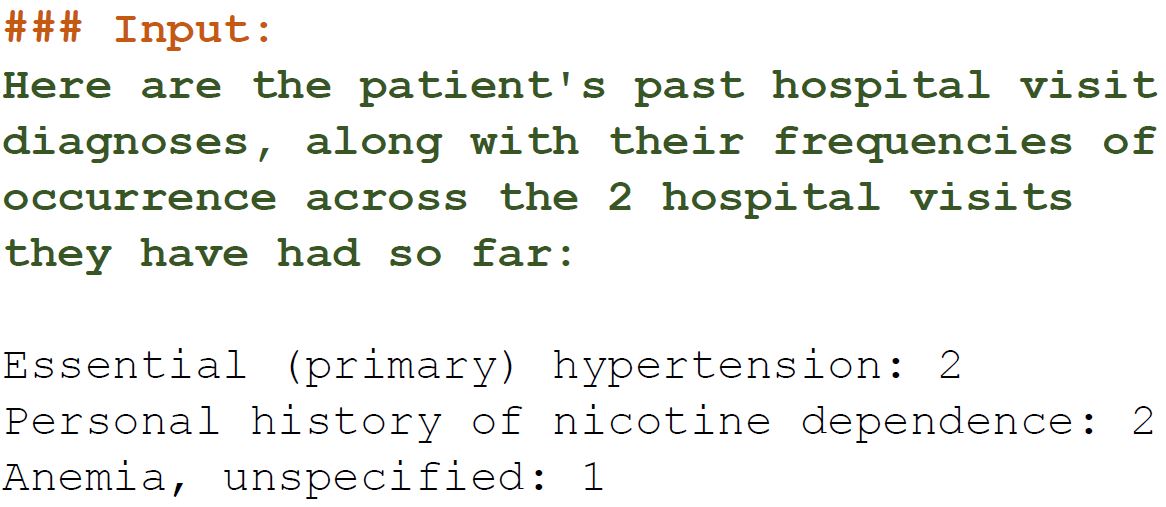}}
  \caption{Input 1}
  \label{fig:input1}
\end{subfigure}%
\begin{subfigure}{.5\textwidth}
  \centering
  \fbox{\includegraphics[width=0.85\linewidth]{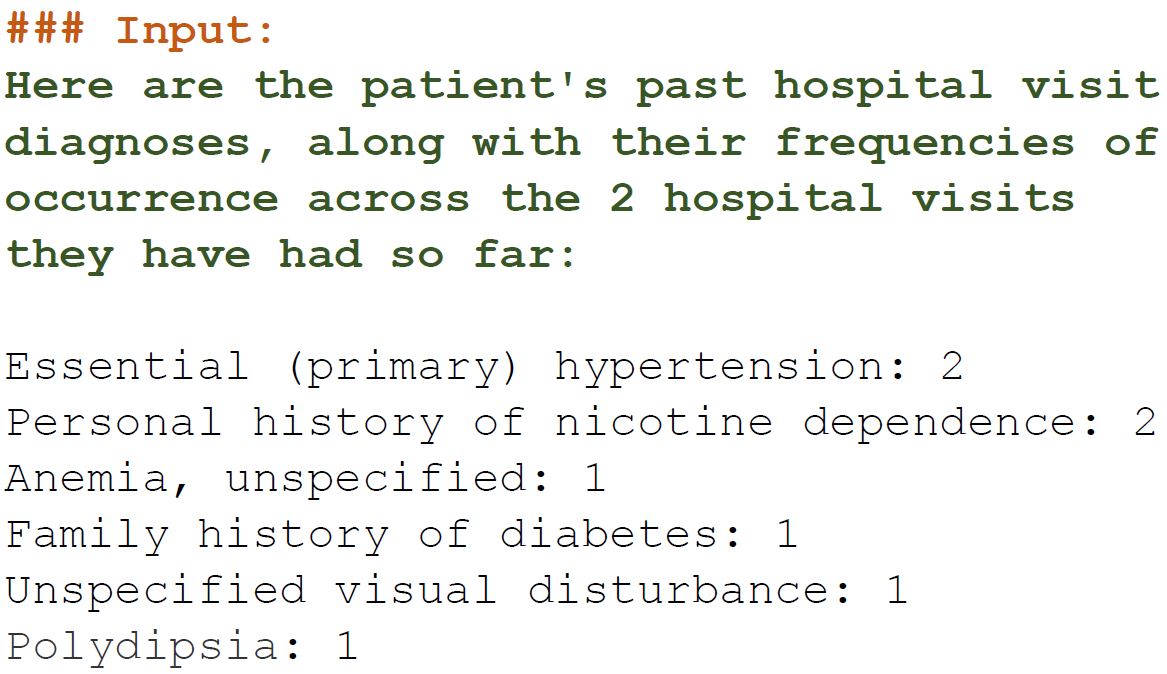}}
  \caption{Input 2}
  \label{fig:input2}
\end{subfigure}
\caption{Examining the variations in LLaMA2-EHR responses when predicting the probability of Diabetes diagnosis based on two distinct inputs: one representing a simple hypothetical patient's medical history (Input 1) and another
involving additional diagnosis information that is known to be the risk factors of Diabetes (Input 2). The objective is to analyze how the likelihood of a ``Yes'' or ``No'' prediction changes within these specific scenarios.}
\label{fig:prompts}
\end{figure*}

We also investigated the sensitivity of the models with respect to changes in the input data. We initially created a simple input containing historical diagnosis information for a hypothetical patient (see Fig. \ref{fig:input1}) and directed the model to predict whether the patient would develop Diabetes in the future using Prompt 1.
Subsequently, we made slight adjustments to the input by including additional past diagnoses recognized as risk factors for Diabetes (as illustrated in Input 2 in Fig. \ref{fig:input2} ). Ideally, we would expect the probability of Yes to increase while doing so. With the MIMIC finetuned model, the probability  from 3\% to 5\%. With the MIMIC-small finetuned model, the probability increased from 2\% to 7\%, a slightly more pronounced change. It is challenging to evaluate these numbers without predetermined benchmarks for the expected increase. Nonetheless, both models did exhibit some level of responsiveness to changes in the input data.

\section{Discussion}
\label{sec:discussion}

\subsection{Summary of  Findings}
In this study, we conducted a comparative examination of various methods for EHR-based risk prediction, with a specific focus on the structured data present in the EHRs. Additionally, we explored two novel approaches (Sent-e-Med and LLaMA2-EHR) for leveraging LMs with these kinds of datasets. To our knowledge, this is the first comprehensive study applying LLMs to structured EHRs for risk prediction.

Overall, our observations indicate that LM-based methods, including MedBERT, Sent-e-Med, and LLaMA2-EHR, consistently outperformed the other three baseline approaches across the majority of the datasets. This trend aligns with prior research findings \cite{gbert, prakash_rarebert_2021, shrestha2022selfsupervised}, highlighting the significant role of pretraining in enhancing performance in downstream tasks, even when working with limited data. In contrast, simpler models like Random Forest and Logistic Regression may not have been able to capture the intricate longitudinal aspects within EHRs.
GRAM, on the other hand, captures longitudinal information in EHRs while providing other benefits such as domain knowledge integration. This model showed comparable performances on Synthea-small but did not perform as well on the other datasets. Deep learning methods typically demand substantial data to acquire effective representations. This requirement for ample data was likely unmet in these datasets. For instance, even in the MIMIC dataset, the number of positive samples is quite low while the vocabulary size remains substantial. In contrast, Synthea-small had a relatively smaller vocabulary size, as evidenced in Table \ref{tab:desc}.

In the context of the SIPPS dataset, all the models being assessed, including simpler ones like Logistic Regression, exhibited nearly indistinguishable performance. However, a more noticeable difference in performance emerged when we assessed the LLaMA2-EHR dataset for Diabetes prediction. The SIPPS dataset mostly consists of outpatient visits and notably features a higher average number of visits (as shown in Table \ref{tab:descr}), which could present challenges in capturing its complexity due to the dataset's limited size.

The MIMIC-small and Synthea-small datasets were primarily introduced to discern whether the performance on SIPPS was influenced by differences in domain or data quality, as opposed to the small sample size. As depicted in Table \ref{tab:results}, the pretraining-based approaches, especially Sent-e-Med and LLaMA2-EHR, outperform the previous approaches on these datasets. This suggests that these approaches perform well even with a limited amount of target data.

Among the LM-based approaches, those that leverage textual descriptions instead of relying solely on medical ontology (Sent-e-Med and LLaMA2-EHR) appear to have harnessed the knowledge encoded within LMs more effectively. 
This proves particularly advantageous when faced with unfamiliar diagnoses during the inference phase or when addressing less commonly encountered medical conditions. 
For instance, in the context of the Synthea-small dataset, specifically within the OUD prediction task, it is evident that Sent-e-Med and LLaMA2-EHR achieved significantly superior performance compared to MedBERT, as indicated by higher PR AUC values.

One plausible rationale for this discrepancy may be attributed to the presence of a substantial proportion (approximately 8\%) of previously unseen diagnosis concepts during the fine-tuning process, referring to codes that were absent in the initial pretraining phase. In contrast to MedBERT's approach of initializing these new codes with random embeddings, these alternative methods draw upon the knowledge contained within their pre-trained representations.
The more pronounced difference in the models' performance within the Synthea-small dataset, particularly in the OUD prediction task, may be explained by the relatively limited number of positive samples available for this specific task.


Notably, LLaMA2-EHR exhibited the most impressive overall performance, which aligns with our expectations, given that LLaMA2 stands as one of the largest LMs currently available and already encompasses a wealth of valuable information that can be advantageous for downstream tasks. The performance difference was more much prominent when looking at the PR AUC values. While ROC AUC values offer insight into the overall separability of data points, PR AUC values place special emphasis on positive examples. When dealing with data imbalances, the examination of both these metrics becomes crucial, as differences between them tend to become more pronounced as shown in Table \ref{tab:results}. The substantially higher PR AUC values achieved by the LLaMA2-EHR model compared to the baseline models suggest that this approach excels even in the presence of data imbalance, as it prioritizes both positive and negative examples effectively. Furthermore, LLaMA2-EHR offers a notably more straightforward approach, especially in comparison to MedBERT and Sent-e-Med, by solely necessitating fine-tuning the EHR corpus.

\subsection{Some Considerations and Future Work}
Considering the preceding discussions and drawing upon prior research \cite{touvron2023llama}, we can deduce LLMs like LLaMA2, demonstrate proficiency in knowledge transfer. Yet, their reliability remains a significant concern, notably due to issues like hallucinations \cite{hallucination}. In our study, we attempted to mitigate these problems by fine-tuning LLaMA2 specifically on the EHR dataset rather than directly using what is available. However, as demonstrated in section \ref{sec:results-table}, fine-tuning led to the model forgetting previously acquired knowledge. Therefore, it's crucial to underline that while these models can yield substantial performance benefits, caution is essential in their application, especially in sensitive tasks such as risk prediction. 

Similarly, the primary task in the study was retrospectively analyzing prior diagnoses to inform future predictions. This choice was driven by our intention in this paper, which was not to optimize the full utilization of available information but to conduct a comparative analysis of variations in a specific data scenario. We recognize that conditions like SUD, OUD, and Diabetes might necessitate a more comprehensive exploration of patient history, including factors like medications, lab results, or social determinants of health for SUD and OUD. Future research could explore the incorporation of such additional data inputs.
It's worth noting that in the case of LLMs, an extended context length may be required to accommodate all the pertinent information, and there are viable solutions available for this purpose \cite{yarn, rope}.

Additionally, while our primary emphasis remained on risk prediction tasks, we did not extensively explore the potential biases that may manifest throughout various phases of machine learning model development. As a part of our future work, we plan to delve deeper into these aspects and gain a more comprehensive understanding of the model predictions.



\bibliographystyle{unsrt}  
\bibliography{references}  

\begin{thebibliography}{10}

\bibitem{pendergrass_using_2018}
Sarah~A. Pendergrass and Dana~C. Crawford.
\newblock Using {Electronic} {Health} {Records} {To} {Generate} {Phenotypes}
  {For} {Research}.
\newblock {\em Current Protocols in Human Genetics}, page e80, December 2018.

\bibitem{risk-prediction}
Benjamin~A Goldstein, Ann~Marie Navar, Michael~J Pencina, and John P~A
  Ioannidis.
\newblock Opportunities and challenges in developing risk prediction models
  with electronic health records data: a systematic review.
\newblock {\em J. Am. Med. Inform. Assoc.}, 24(1):198--208, January 2017.

\bibitem{gram}
Edward Choi, Mohammad~Taha Bahadori, Le~Song, Walter~F. Stewart, and Jimeng
  Sun.
\newblock Gram: Graph-based attention model for healthcare representation
  learning.
\newblock In {\em Proceedings of the 23rd ACM SIGKDD International Conference
  on Knowledge Discovery and Data Mining}, KDD '17, page 787–795. Association
  for Computing Machinery, 2017.

\bibitem{gbert}
Junyuan Shang, Tengfei Ma, Cao Xiao, and Jimeng Sun.
\newblock Pre-training of graph augmented transformers for medication
  recommendation.
\newblock {\em CoRR}, abs/1906.00346, 2019.

\bibitem{icd-10}
Icd-10: History and context.

\bibitem{rasmy_med-bert_2021}
Laila Rasmy, Yang Xiang, Ziqian Xie, Cui Tao, and Degui Zhi.
\newblock Med-{BERT}: pretrained contextualized embeddings on large-scale
  structured electronic health records for disease prediction.
\newblock {\em npj Digital Medicine}, 4(1):86, December 2021.

\bibitem{prakash_rarebert_2021}
Pks Prakash, Srinivas Chilukuri, Nikhil Ranade, and Shankar Viswanathan.
\newblock {RareBERT}: {Transformer} {Architecture} for {Rare} {Disease}
  {Patient} {Identification} using {Administrative} {Claims}.
\newblock {\em Proceedings of the AAAI Conference on Artificial Intelligence},
  35:453--460, May 2021.

\bibitem{devlin_bert_2019}
Jacob Devlin, Ming-Wei Chang, Kenton Lee, and Kristina Toutanova.
\newblock {BERT}: {Pre}-training of {Deep} {Bidirectional} {Transformers} for
  {Language} {Understanding}, May 2019.

\bibitem{li_behrt_2020}
Yikuan Li, Shishir Rao, José Roberto~Ayala Solares, Abdelaali Hassaine, Rema
  Ramakrishnan, Dexter Canoy, Yajie Zhu, Kazem Rahimi, and Gholamreza
  Salimi-Khorshidi.
\newblock {BEHRT}: {Transformer} for {Electronic} {Health} {Records}.
\newblock {\em Scientific Reports}, 10:7155, December 2020.

\bibitem{economic_cost_OUD}
Feijun Luo.
\newblock State-{Level} {Economic} {Costs} of {Opioid} {Use} {Disorder} and
  {Fatal} {Opioid} {Overdose} — {United} {States}, 2017.
\newblock {\em MMWR. Morbidity and Mortality Weekly Report}, 70, 2021.

\bibitem{gensyn}
Angeela Acharya, Siddhartha Sikdar, Sanmay Das, and Huzefa Rangwala.
\newblock Gensyn: A multi-stage framework for generating synthetic microdata
  using macro data sources.
\newblock In {\em 2022 IEEE International Conference on Big Data (Big Data)},
  pages 685--692, 2022.

\bibitem{county-level}
Angeela Acharya, Alyssa~M. Izquierdo, Stefanie~F. Gonçalves, Rebecca~A. Bates,
  Faye~S. Taxman, Martin~P. Slawski, Huzefa~S. Rangwala, and Siddhartha Sikdar.
\newblock Exploring county-level spatio-temporal patterns in opioid overdose
  related emergency department visits.
\newblock {\em PLOS ONE}, 17(12):1--15, 12 2022.

\bibitem{McLellan2017-eu}
A~Thomas McLellan.
\newblock Substance misuse and substance use disorders: Why do they matter in
  healthcare?
\newblock {\em Trans. Am. Clin. Climatol. Assoc.}, 128:112--130, 2017.

\bibitem{touvron2023llama}
Hugo Touvron, Thibaut Lavril, Gautier Izacard, Xavier Martinet, Marie-Anne
  Lachaux, Timothée Lacroix, Baptiste Rozière, Naman Goyal, Eric Hambro,
  Faisal Azhar, Aurelien Rodriguez, Armand Joulin, Edouard Grave, and Guillaume
  Lample.
\newblock Llama: Open and efficient foundation language models, 2023.

\bibitem{palm}
Aakanksha Chowdhery, Sharan Narang, Jacob Devlin, Maarten Bosma, Gaurav Mishra,
  Adam Roberts, Paul Barham, Hyung~Won Chung, Charles Sutton, Sebastian
  Gehrmann, Parker Schuh, Kensen Shi, Sasha Tsvyashchenko, Joshua Maynez,
  Abhishek Rao, Parker Barnes, Yi~Tay, Noam Shazeer, Vinodkumar Prabhakaran,
  Emily Reif, Nan Du, Ben Hutchinson, Reiner Pope, James Bradbury, Jacob
  Austin, Michael Isard, Guy Gur-Ari, Pengcheng Yin, Toju Duke, Anselm
  Levskaya, Sanjay Ghemawat, Sunipa Dev, Henryk Michalewski, Xavier Garcia,
  Vedant Misra, Kevin Robinson, Liam Fedus, Denny Zhou, Daphne Ippolito, David
  Luan, Hyeontaek Lim, Barret Zoph, Alexander Spiridonov, Ryan Sepassi, David
  Dohan, Shivani Agrawal, Mark Omernick, Andrew~M. Dai,
  Thanumalayan~Sankaranarayana Pillai, Marie Pellat, Aitor Lewkowycz, Erica
  Moreira, Rewon Child, Oleksandr Polozov, Katherine Lee, Zongwei Zhou, Xuezhi
  Wang, Brennan Saeta, Mark Diaz, Orhan Firat, Michele Catasta, Jason Wei,
  Kathy Meier-Hellstern, Douglas Eck, Jeff Dean, Slav Petrov, and Noah Fiedel.
\newblock Palm: Scaling language modeling with pathways, 2022.

\bibitem{gpt3}
Tom Brown, Benjamin Mann, Nick Ryder, Melanie Subbiah, Jared~D Kaplan, Prafulla
  Dhariwal, Arvind Neelakantan, Pranav Shyam, Girish Sastry, Amanda Askell,
  Sandhini Agarwal, Ariel Herbert-Voss, Gretchen Krueger, Tom Henighan, Rewon
  Child, Aditya Ramesh, Daniel Ziegler, Jeffrey Wu, Clemens Winter, Chris
  Hesse, Mark Chen, Eric Sigler, Mateusz Litwin, Scott Gray, Benjamin Chess,
  Jack Clark, Christopher Berner, Sam McCandlish, Alec Radford, Ilya Sutskever,
  and Dario Amodei.
\newblock Language models are few-shot learners.
\newblock In H.~Larochelle, M.~Ranzato, R.~Hadsell, M.F. Balcan, and H.~Lin,
  editors, {\em Advances in Neural Information Processing Systems}, volume~33,
  pages 1877--1901. Curran Associates, Inc., 2020.

\bibitem{cpt}
Cpt® codes: What are they, why are they necessary, and how are they
  developed?, October 2023.

\bibitem{rxnorm}
Normalized names for clinical drugs: Rxnorm at 6 years, July 2011.

\bibitem{opiod-prediction}
X~Dong, S~Rashidian, Y~Wang, J~Hajagos, X~Zhao, R~N Rosenthal, J~Kong, M~Saltz,
  J~Saltz, and F~Wang.
\newblock Machine learning based opioid overdose prediction using electronic
  health records. {AMIA}.
\newblock In {\em Annual Symposium proceedings. {AMIA} Symposium}, pages
  389--398. 2019.

\bibitem{choi_doctor_2016}
Edward Choi, Mohammad~Taha Bahadori, Andy Schuetz, Walter~F. Stewart, and
  Jimeng Sun.
\newblock Doctor {AI}: {Predicting} {Clinical} {Events} via {Recurrent}
  {Neural} {Networks}.
\newblock {\em JMLR workshop and conference proceedings}, 56:301--318, August
  2016.

\bibitem{Zhang_2018}
Jinghe Zhang, Kamran Kowsari, James~H. Harrison, Jennifer~M. Lobo, and Laura~E.
  Barnes.
\newblock Patient2vec: A personalized interpretable deep representation of the
  longitudinal electronic health record.
\newblock {\em IEEE Access}, 2018.

\bibitem{DAI2015189}
Wuyang Dai, Theodora~S. Brisimi, William~G. Adams, Theofanie Mela, Venkatesh
  Saligrama, and Ioannis~Ch. Paschalidis.
\newblock Prediction of hospitalization due to heart diseases by supervised
  learning methods.
\newblock {\em International Journal of Medical Informatics}, 2015.

\bibitem{random-forest}
Sreekanth Rallapalli and T.~Suryakanthi.
\newblock Predicting the risk of diabetes in big data electronic health records
  by using scalable random forest classification algorithm.
\newblock In {\em 2016 International Conference on Advances in Computing and
  Communication Engineering (ICACCE)}, pages 281--284, 2016.

\bibitem{retain}
Edward Choi, Mohammad~Taha Bahadori, Joshua~A. Kulas, Andy Schuetz, Walter~F.
  Stewart, and Jimeng Sun.
\newblock Retain: An interpretable predictive model for healthcare using
  reverse time attention mechanism.
\newblock In {\em Proceedings of the 30th International Conference on Neural
  Information Processing Systems}, NIPS'16, page 3512–3520. Curran Associates
  Inc., 2016.

\bibitem{Barbieri2020BenchmarkingDL}
Sebastiano Barbieri, James~P. Kemp, Oscar Perez-Concha, Sradha~S Kotwal, Martin
  Gallagher, Angus Ritchie, and Louisa Jorm.
\newblock Benchmarking deep learning architectures for predicting readmission
  to the icu and describing patients-at-risk.
\newblock {\em Scientific Reports}, 2020.

\bibitem{mimiciv_v1}
Alistair Johnson, Lucas Bulgarelli, Tom Pollard, Steven Horng, Leo~Anthony
  Celi, and Roger Mark.
\newblock Mimic-iv, 2021.

\bibitem{pollard_eicu_2018}
Tom~J. Pollard, Alistair E.~W. Johnson, Jesse~D. Raffa, Leo~A. Celi, Roger~G.
  Mark, and Omar Badawi.
\newblock The {eICU} {Collaborative} {Research} {Database}, a freely available
  multi-center database for critical care research.
\newblock {\em Scientific Data}, 5:180178, September 2018.

\bibitem{icd-heirarchy}
Elias Moons, Aditya Khanna, Abbas Akkasi, and Marie-Francine Moens.
\newblock A comparison of deep learning methods for icd coding of clinical
  records.
\newblock {\em Applied Sciences}, 10(15), 2020.

\bibitem{10.1007/978-3-030-61255-9_14}
Angeela Acharya, Jitin Krishnan, Desmond Arias, and Huzefa Rangwala.
\newblock Homicidal event forecasting and interpretable analysis using
  hierarchical attention model.
\newblock In Robert Thomson, Halil Bisgin, Christopher Dancy, Ayaz Hyder, and
  Muhammad Hussain, editors, {\em Social, Cultural, and Behavioral Modeling},
  pages 140--150, Cham, 2020. Springer International Publishing.

\bibitem{cehr-bert}
Chao Pang, Xinzhuo Jiang, Krishna~S. Kalluri, Matthew Spotnitz, RuiJun Chen,
  Adler Perotte, and Karthik Natarajan.
\newblock Cehr-bert: Incorporating temporal information from structured ehr
  data to improve prediction tasks.
\newblock In Subhrajit Roy, Stephen Pfohl, Emma Rocheteau, Girmaw~Abebe
  Tadesse, Luis Oala, Fabian Falck, Yuyin Zhou, Liyue Shen, Ghada Zamzmi,
  Purity Mugambi, Ayah Zirikly, Matthew B.~A. McDermott, and Emily Alsentzer,
  editors, {\em Proceedings of Machine Learning for Health}, volume 158 of {\em
  Proceedings of Machine Learning Research}, pages 239--260. PMLR, 04 Dec 2021.

\bibitem{clinical-bert}
Kexin Huang, Jaan Altosaar, and Rajesh Ranganath.
\newblock Clinicalbert: Modeling clinical notes and predicting hospital
  readmission.
\newblock {\em CoRR}, abs/1904.05342, 2019.

\bibitem{yang2022gatortron}
Xi~Yang, Aokun Chen, Nima PourNejatian, Hoo~Chang Shin, Kaleb~E Smith,
  Christopher Parisien, Colin Compas, Cheryl Martin, Mona~G Flores, Ying Zhang,
  Tanja Magoc, Christopher~A Harle, Gloria Lipori, Duane~A Mitchell, William~R
  Hogan, Elizabeth~A Shenkman, Jiang Bian, and Yonghui Wu.
\newblock Gatortron: A large clinical language model to unlock patient
  information from unstructured electronic health records, 2022.

\bibitem{snomed_ct}
Christophe Gaudet-Blavignac, Vasiliki Foufi, Mina Bjelogrlic, and Christian
  Lovis.
\newblock Use of the systematized nomenclature of medicine clinical terms
  (snomed ct) for processing free text in health care: Systematic scoping
  review.
\newblock {\em J Med Internet Res}, Jan 2021.

\bibitem{johnson_alistair_mimic-iv_nodate}
Alistair Johnson, Lucas Bulgarelli, Tom Pollard, Steven Horng, Leo~Anthony
  Celi, and Roger Mark.
\newblock {MIMIC}-{IV}.

\bibitem{synthea}
Jason Walonoski, Mark Kramer, Joseph Nichols, Andre Quina, Chris Moesel, Dylan
  Hall, Carlton Duffett, Kudakwashe Dube, Thomas Gallagher, and Scott
  McLachlan.
\newblock {Synthea: An approach, method, and software mechanism for generating
  synthetic patients and the synthetic electronic health care record}.
\newblock {\em Journal of the American Medical Informatics Association}, 2017.

\bibitem{deep_oud}
Aditya Kashyap, Chris Callison-Burch, and Mary~Regina Boland.
\newblock A deep learning method to detect opioid prescription and opioid use
  disorder from electronic health records.
\newblock {\em International Journal of Medical Informatics}, 171:104979, 2023.

\bibitem{deep_ehr}
Benjamin Shickel, Patrick Tighe, Azra Bihorac, and Parisa Rashidi.
\newblock Deep ehr: A survey of recent advances on deep learning techniques for
  electronic health record (ehr) analysis.
\newblock {\em CoRR}, abs/1706.03446, 2017.

\bibitem{reimers_sentence-bert_2019}
Nils Reimers and Iryna Gurevych.
\newblock Sentence-{BERT}: {Sentence} {Embeddings} using {Siamese}
  {BERT}-{Networks}.
\newblock 2019.
\newblock Publisher: arXiv Version Number: 1.

\bibitem{ferrari_triplet_2018}
Xingping Dong and Jianbing Shen.
\newblock Triplet {Loss} in {Siamese} {Network} for {Object} {Tracking}.
\newblock In {\em Computer {Vision} – {ECCV} 2018}, volume 11217, pages
  472--488. Springer International Publishing, 2018.

\bibitem{cer_universal_2018}
Daniel Cer, Yinfei Yang, Sheng-yi Kong, Nan Hua, Nicole Limtiaco, Rhomni~St.
  John, Noah Constant, Mario Guajardo-Cespedes, Steve Yuan, Chris Tar,
  Yun-Hsuan Sung, Brian Strope, and Ray Kurzweil.
\newblock Universal {Sentence} {Encoder}.
\newblock 2018.
\newblock Publisher: arXiv Version Number: 2.

\bibitem{roberta}
Yinhan Liu, Myle Ott, Naman Goyal, Jingfei Du, Mandar Joshi, Danqi Chen, Omer
  Levy, Mike Lewis, Luke Zettlemoyer, and Veselin Stoyanov.
\newblock Roberta: {A} robustly optimized {BERT} pretraining approach.
\newblock {\em CoRR}, abs/1907.11692, 2019.

\bibitem{LLM}
Humza Naveed, Asad~Ullah Khan, Shi Qiu, Muhammad Saqib, Saeed Anwar, Muhammad
  Usman, Naveed Akhtar, Nick Barnes, and Ajmal Mian.
\newblock A comprehensive overview of large language models, 2023.

\bibitem{catastrophic}
Prakhar Kaushik, Alex Gain, Adam Kortylewski, and Alan~L. Yuille.
\newblock Understanding catastrophic forgetting and remembering in continual
  learning with optimal relevance mapping.
\newblock {\em CoRR}, abs/2102.11343, 2021.

\bibitem{shrestha2022selfsupervised}
Sulabh Shrestha, Yimeng Li, and Jana Kosecka.
\newblock Self-supervised pre-training for semantic segmentation in an indoor
  scene, 2022.

\bibitem{hallucination}
Nick McKenna, Tianyi Li, Liang Cheng, Mohammad~Javad Hosseini, Mark Johnson,
  and Mark Steedman.
\newblock Sources of hallucination by large language models on inference tasks,
  2023.

\bibitem{yarn}
Bowen Peng, Jeffrey Quesnelle, Honglu Fan, and Enrico Shippole.
\newblock Yarn: Efficient context window extension of large language models,
  2023.

\bibitem{rope}
Shouyuan Chen, Sherman Wong, Liangjian Chen, and Yuandong Tian.
\newblock Extending context window of large language models via positional
  interpolation, 2023.

\bibitem{focal_loss}
Tsung-Yi Lin, Priya Goyal, Ross Girshick, Kaiming He, and Piotr Dollár.
\newblock Focal loss for dense object detection.
\newblock In {\em 2017 IEEE International Conference on Computer Vision
  (ICCV)}, pages 2999--3007, 2017.

\end{thebibliography}

\section{Supplementary Materials}
\subsection{Sub-group Distribution}
Table \ref{tab:desc} offers a breakdown of different subgroups within the MIMIC dataset. It's worth noting that a single patient may have multiple statuses recorded throughout their medical history (e.g., transitioning from married to divorced). However, it is observed that the distribution remains relatively consistent whether we consider the first or last recorded attribute.
Here, we have reported the statistics based on the initial recorded attribute for each patient. 

\begin{table*}[h]
\footnotesize
\begin{center}
\caption{Exploratory analysis of patient attributes in MIMIC: The table presents the percentage of patients with and without condition (OUD/SUD/Diabetes) in various demographic groups.}
\label{tab:desc}
\begin{tabular}{ccccccc}

\hline
\textbf{Attribute} & \textbf{SUD (\%)} & \textbf{No SUD (\%)} & \textbf{OUD (\%)} & \textbf{No OUD (\%)} & \textbf{Diabetes (\%)} & \textbf{No Diabetes (\%)} \\ 
\hline
\textbf{Insurance }& ~ & ~ & ~ & ~ \\ 
Medicare & 0.32 & 0.37 & 0.29 & 0.36&0.44&0.32 \\ 
Medicaid & 0.16 & 0.07 & 0.26 & 0.08&0.09&0.09 \\ 
Others & 0.52 & 0.56 & 0.45 & 0.56 &0.47& 0.59 \\ 
\hline
\textbf{Language} & ~ & ~ & ~ & ~ \\ 
English & 0.95 & 0.89 & 0.97 & 0.90 &0.88&0.92 \\ 
Others & 0.05 & 0.11 & 0.03 & 0.10 &0.12&0.08\\ 
\hline
\textbf{Marital Status} & ~ & ~ & ~ & ~ \\ 
Single & 0.52 & 0.31 & 0.64 & 0.34&0.33&0.37 \\ 
Married & 0.30 & 0.48 & 0.19 & 0.45 &0.44&0.44\\ 
Divorced & 0.09 & 0.06 & 0.10 & 0.07 &0.08&0.06\\ 
Widowed & 0.06 & 0.12 & 0.04 & 0.11 &0.12&0.1\\ 
\hline
\textbf{Ethnicity }& ~ & ~ & ~ & ~ \\ 
White & 0.67 & 0.68 & 0.75 & 0.68 &0.62&0.7\\ 
Black/African American & 0.19 & 0.14 & 0.14 & 0.15&0.22&0.13 \\ 
Hispanic/ Latino & 0.06 & 0.05 & 0.06 & 0.05&0.07&0.05 \\ 
Other & 0.04 & 0.05 & 0.03 & 0.05&0.05&0.04 \\ 
Asian & 0.01 & 0.04 & 0.00 & 0.04 &0.03&0.04\\ 
American Indian/Alaskan native & 0.00 & 0.00 & 0.00 & 0.00 &0.001&0.002 \\ 
Unknown & 0.02 & 0.03 & 0.02 & 0.03 &0.02&0.03\\ 
\hline
\end{tabular}
\end{center}
\end{table*}

\subsection{Implementation Details of Sent-e-Med}
Given the relatively smaller size of the data used for both pretraining and fine-tuning, we constrained the number of parameters in our model in comparison to the original BERT model. For the transformer layer, we used 4 hidden layers, 4 attention heads, and a hidden dimension of 384 (matching the size of the embedding returned by SBERT). The \replaced[id=ss]{dimension of the linear layer}{feedforward size} was set to 64. The maximum sequence length of 128 tokens (sentences) was used since we do not work with long sequences. For optimization, the AdamWeight decay optimizer was used with a learning rate of 1e-5. We used a 32 GB V100 GPU. The pretraining took around 5 days while the finetuning took only around 20 minutes.

For the risk prediction tasks, we employed the binary cross-entropy loss function. Given the highly imbalanced nature of our data, we also explored alternative weight-balanced loss functions, including the focal loss \cite{focal_loss}. Interestingly, none of these alternative loss functions resulted in improved model performance. In fact, the cross-entropy loss consistently yielded the most stable outcomes across our datasets. 

\subsection{Implementation Details of LLaMA2-EHR}
LLaMA2-EHR was finetuned for a total of 5 epochs and the best checkpoint was selected for inference. Each model was trained on a single 80GB A100 GPU. Fine-tuning required approximately two days for MIMIC and approximately six hours for Synthea, SIPPS, and MIMIC-small datasets.

\begin{figure}[!t]
\centering
\includegraphics[width=0.65\textwidth]{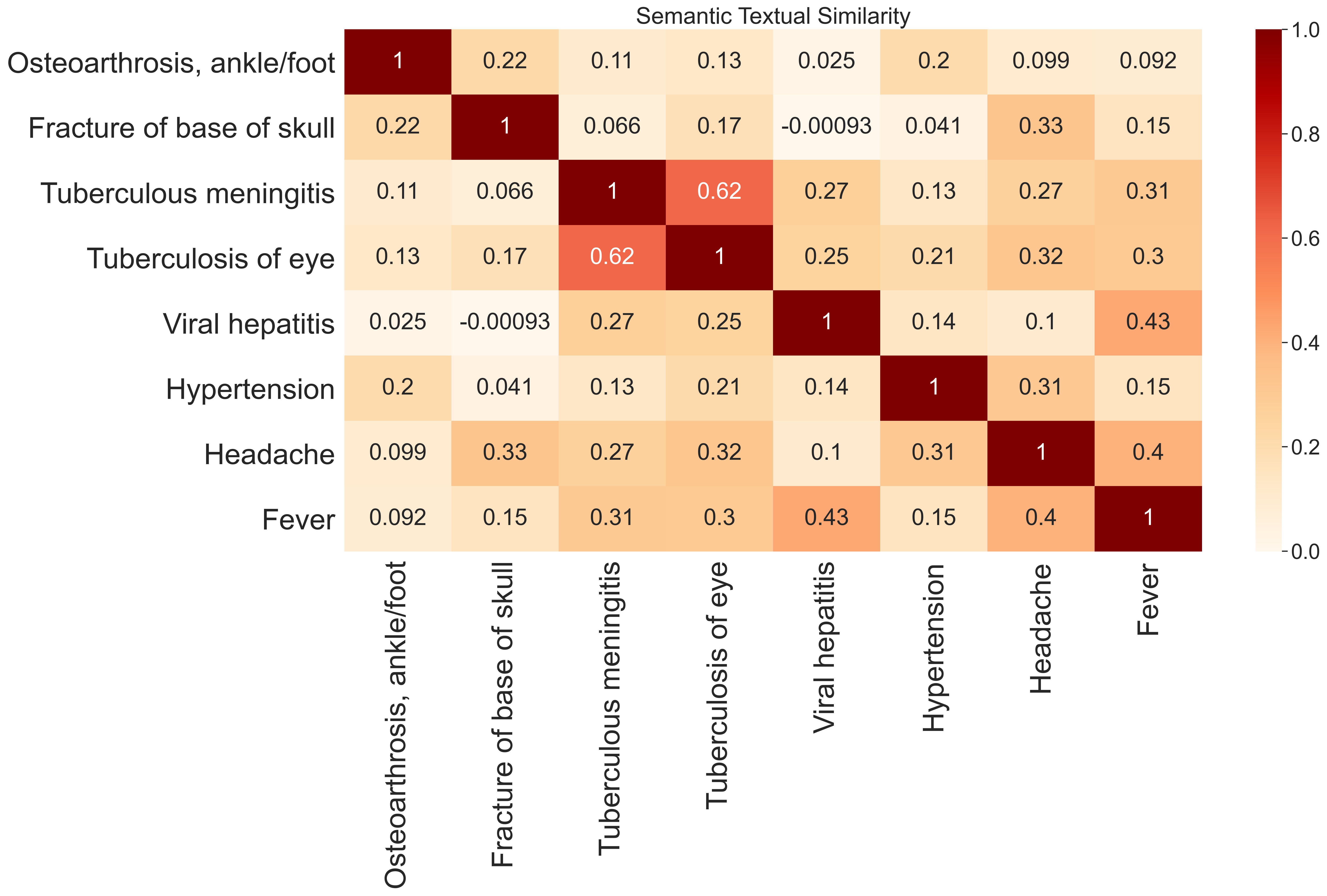}
\caption{Heatmap of the cosine similarities between the descriptions of the medical codes}
\label{fig:similarity}
\end{figure}

\subsection{Sentence Encoding Outputs}
The embeddings corresponding to the diagnosis descriptions returned by SBERT can be assessed based on cosine similarity to see how they relate to each other. \added[id=ss]{Pairs of diagnoses which are very much related to each other should have high cosine similarity (closer to 1) with each other and conversely, unrelated ones should have lower cosine similarity (closer to 0).} Fig \ref{fig:similarity} represents a heatmap of the cosine similarity scores between the textual descriptions of some of the common diagnoses. \replaced[id=ss]{From the figure, we can see that as desired,}{As shown,} the diagnoses that are related to each other have high cosine similarity scores. For instance, ``Tuberculosis meningitis'' and ``Tuberculosis of eye'' have a cosine similarity score of 0.62, which is expected because they are related conditions. ``Tuberculosis meningitis'' and ``Viral hepatitis'' have a cosine similarity score of 0.27 because they are not entirely unrelated in that they are both infectious diseases. On the other hand, ``Tuberculosis meningitis'' and ``Fracture of base of skull'' have a very low similarity score (0.066) as they are extremely unlikely to be related. \deleted[id=ss]{That is expected because they are completely unrelated conditions.} This shows that the model is able to capture the semantic relationships between two diagnoses and encodes them in such a way that related diagnoses are close to each other in cosine distance thus providing verification to our motivation for using sentence embeddings to encode the medical codes.  
\end{document}